 \documentclass[smallabstract,smallcaptions]{dccpaper}

\usepackage{epsfig}
\usepackage{amsmath}
\usepackage{amssymb}
\usepackage{color}
\usepackage{url}
\usepackage{svg}
\usepackage{array}
\usepackage{multirow}
\usepackage{tabularx}
\usepackage{algorithm}
\usepackage{algpseudocode}
\usepackage{graphicx}
\newlength{\figurewidth}
\newlength{\smallfigurewidth}

\begin{document}

\title
{\large
\textbf{Complexity Reduction Study Based on RD Costs Approximation for VVC Intra Partitioning}
}

\author{%
 M.E.A. Kherchouche$^{\ast \dag}$, F. Galpin$^{\ast}$, T. Dumas$^{\ast}$, F. Schnitzler$^{\ast}$, \\D. Menard$^{\dag}$, L. Zhang$^{\dag}$\\[0.5em]
{\small\begin{minipage}{\linewidth}\begin{center}
\begin{tabular}{ccc}
$^{\ast}$InterDigital, R\&I &&  $^{\dag}$Univ Rennes, INSA Rennes\\
845a Avenue des Champs Blancs && CNRS, IETR - UMR 6164 \\
35510 Cesson-Sevigne &&  F-35000 Rennes, France\\
\url{firstname.lastname@interdigital.com} && \url{firstname.lastname@insa-rennes.fr}
\end{tabular}
\end{center}\end{minipage}}
}

\maketitle
\thispagestyle{empty}

\begin{abstract}
In this paper, a complexity study is conducted for Versatile Video Codec (VVC) intra partitioning to accelerate the exhaustive search involved in Rate-Distortion Optimization (RDO) process. To address this problem, two main machine learning techniques are proposed and compared. Unlike existing methods, the proposed approaches are size independent and incorporate the Rate-Distortion (RD) costs of neighboring blocks as input features. The first method is a regression based technique that predicts normalized RD costs of a given Coding Unit (CU). As partitioning possesses the Markov property, the associated decision-making problem can be modeled as a Markov Decision Process (MDP) and solved by Reinforcement Learning (RL). The second approach is a RL agent learned from trajectories of CU decision across two depths with Deep Q-Network (DQN) algorithm. Then a pre-determined thresholds are applied for both methods to select a suitable split for the current CU.  

\end{abstract}

\Section{Introduction}
The VVC \cite{VVC} standard aims to improve encoding efficiency about $50\%$ but in return a considerable  increase in complexity for the encoding process compared to High Efficiency Video Codec (HEVC)  \cite{HEVC}.  The encoding process can be designed as a combinatorial optimization problem, comprising multiple stages such as block partitioning, intra/inter prediction, transformation, quantization, and entropy coding. This study focuses on the acceleration of the partitioning part in intra coded slice in the encoder side. In HEVC, partitioning is limited to a Quad Tree (QT) structure, where a $64\times64$ Coding Tree Unit (CTU) is recursively divided into four equal-sized, non-overlapping blocks. VVC enhances this approach by incorporating rectangular partitions through the Multi-Type Tree (MTT) tool, which includes Binary Tree (BT) and Ternary Tree (TT) splits. BT divides a CU into two equal rectangular blocks, either vertically (BTV) or horizontally (BTH), while TT partitions the CU into three rectangular blocks with a $1:2:1$ size ratio, applicable in both vertical (TTV) and horizontal (TTH) directions. The partitioning scheme is determined by the RDO process, which balances the trade-off between the quality of the reconstructed region and the bit rate required for transmission. Conventional encoder heuristics, typically based on analyzing a large set of encoding, reduce the need for exhaustive searches across all possible CU partitioning combinations. However, such heuristics are usually based on early decision based on current RD cost, without taking into account the content itself. Neural Networks (NNs), trained via deep learning algorithms, offer a complementary approach to address the combinatorial complexity and can be integrated into the encoder to accelerate the RDO process efficiently.

Previous works have presented some fast algorithms for reducing the partitioning complexity of HEVC and VVC. Indeed, evaluating a partition mode by performing the encoding to obtain the RD cost is computationally expensive. For that, two main categories can be considered: The first category aims to skip unnecessary splitting modes based on supervised classification methods.  Li \textit{et al.} \cite{DeepQTMT} propose a Multi-Stage Exit Convolutional Neural Network (MSE-CNN) combined with an early-exit mechanism to determine the CU partition. Tissier \textit{et al.} \cite{Tissier2023} propose a two-stage learning-based technique to tackle the complexity overhead of MTT in VVC intra encoders. Na Li \textit{et al.} \cite{ACTCRTHEVC} propose an early termination approach of CU splitting based on actor-critic algorithm to learn classifiers for low complexity video coding.
The second category includes approaches that estimate the cost of the possible splitting modes. In \cite{DQNHEVC}, Chung Chia-Hua \textit{et al.} introduce a Q-Learning algorithm to learn a CNN for HEVC to approximate the RD cost reduction of each possible state-action pair.  In \cite{RL_VVC}, Zhao Jinchao \textit{et al.}  propose a fast CU splitting decision method based on Deep Reinforcement Learning (DRL) for VVC to decrease the coding complexity for $32\times32$ CUs.  In our prior work \cite{Reg}, a CNN is employed to process a $32\times32$ CU of the luma component along with the associated Quantization Parameter (QP) as inputs. The model outputs a vector of six RD costs, one for each split mode, and only a subset of these modes (up to 6) is tested based on pre-defined thresholds.

This paper presents a new complexity reduction study comparing two main techniques, both classified under the second category. The first approach is RD costs regression based NN intra partitioning for $32\times32, 16\times16$ and $8\times8$ CUs. The second technique is a DRL algorithm that learns an agent to approximate RD costs for $32\times32$  and $16\times16$. Unlike state of the art works, both proposed methods are size independent and can be a single NN that treats several CU sizes.

The paper is organized as follows. Section \ref{sec:VTMQT} explains the fixed configuration used as a benchmark for this study, CUs representation and the considered outputs in section \ref{sec:IO}, sections \ref{sec:Reg}, \ref{sec:RL}
present the overview of the two approaches, the results are in section  \ref{sec:Exp} and finally section \ref{sec:Conc} concludes the paper.
\section{VTM Fixed Configuration}
\label{sec:VTMQT}

\begin{figure}[ht]
    \centering
    \begin{minipage}{0.5\textwidth}
        \centering
        \includegraphics[height=5cm, width=\textwidth]{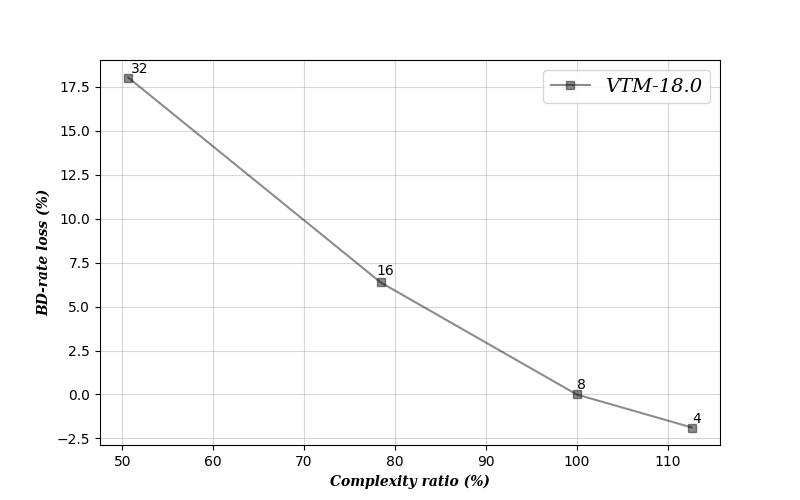}
    \end{minipage}\hfill
    \begin{minipage}{0.5\textwidth}
        \centering
        \includegraphics[height=5cm,width=\textwidth]{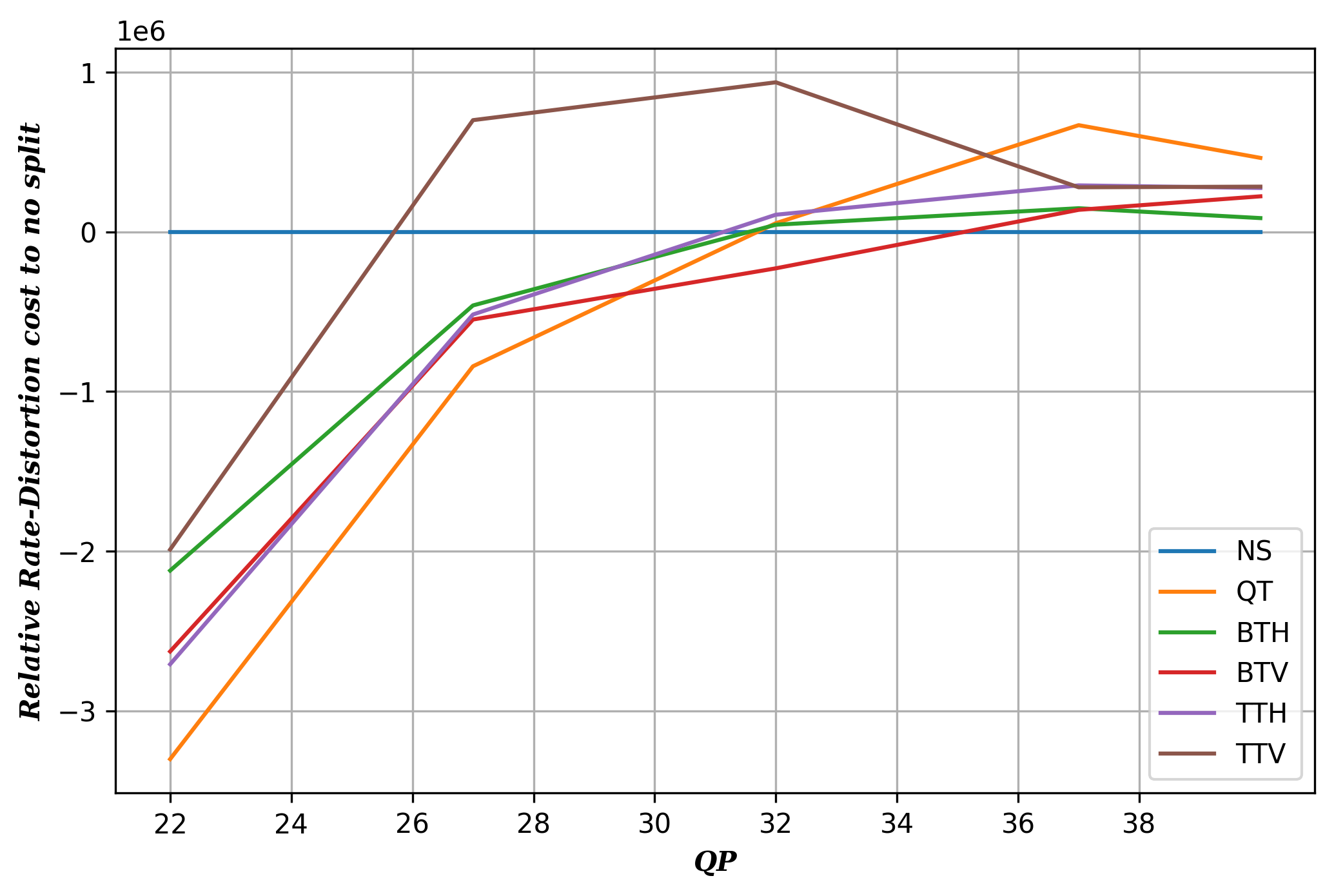}
    \end{minipage}
    \caption{\textbf{Left:} Benchmark VTM-18.0 in All Intra configuration with four QTDepths. \textbf{Right:} Rate-distortion costs of a $32\times32$ CU for each splitting mode at different quantization parameter values. NS RD cost is subtracted from all RD costs.}
\label{fig:VTM-18.0}
\end{figure}

The VVC Test Model (VTM) serves as the reference software for implementing and testing coding tools defined by the VVC standard, with VTM-18.0 used in All Intra (AI) configuration for this study. The initial focus is on Quad Tree (QT) partitioning due to its high computational cost and better control over the solutions, with the potential for future extension to MTT. Each CU may either split using QT or remain unsplit so-called No Split (NS). Figure \ref{fig:VTM-18.0} (Left) shows the complexity versus bd-rate trade-offs of VTM-18.0 at various Quad Tree Depths (QTDepth), where QTDepth indicates the level of recursive partitioning. The figure highlights that a QTDepth of $3$ corresponds to a minimum CU block size of $8\times8$, with similar interpretations for QTDepths $1, 2, 4$ and their respective minimum block sizes of $32\times32$, $16\times16$, and $4\times4$. Complexity reduction is defined as the difference in the total number of pixels processed during encoding between a specified QTDepth and the default QTDepth of $3$, making it independent of VTM implementations and is calculated as follows:

{\small
\begin{equation}
    \Delta C = \frac{1}{4} \sum_{QP_i \in \{22,27,32,37\}}{\frac{T_3(QP_i) - T_dQP_i)}{T_3(QP_i)}}
\end{equation}
}
Where $T_3(QP_i)$ is the complexity at QTDepth 3 and  $T_d(QP_i)$ refers to the complexity at QTDepth $d$ where $d \in [1, 4]$.

\section{CU vector representation and RD costs approximation}
\label{sec:IO}
Since a NN takes a fixed input size, the CU representation is designed to be a vector of fixed dimension that includes four feature sets including: neighbor information, parent information, block information and spatial information based on features extracted from the pixels of the current block. This method makes the two solutions considered in this paper size independent and only one NN can handle several CU sizes.

\subsection{Neighbor Information (NI)}
The encoding process follows a raster scanning order, making relevant information from previously encoded blocks available for the current CU. The RD costs of the top and left CUs are used as neighboring features and are normalized per pixel by dividing by the block size. Additionally, the QTDepth of the top and left blocks are extracted and normalized by the maximum QTDepth, which is 4.

\subsection{Parent Information (PI)}
During the RDO process, a CU generates sub-CUs, meaning each CU has a parent CU. To capture the partitioning structure, it may be important to include the split mode costs of the parent CU as informative data for the child CU. Since only QT configuration is allowed, the NS RD cost is considered as a parent feature, normalized per pixel. The RD cost, $J = R + \lambda . D,$
where $R$ is the rate, $D$ is the distortion and $\lambda$ to balance the two terms. 
Additionally to the parent NS cost, normalized rate and distortion are included separately as parent features.

\subsection{Block Information (BI)}
The feature set includes the height and width of the current CU, normalized by the maximum CTU size in VVC (128), and the QP, normalized by 64. In this study, intra mode is enabled in the RDO, and the recursive QT is controlled by the proposed methods. Consequently, the normalized NS cost of the current CU is also considered as part of the feature set.

\subsection{Spatial Information (SI)}
The RDO process in AI configuration relies on spatial information. To extract spatial features from a CU, four rows and columns of causal pixels on the top and left are added. The CU is then divided into four sub-blocks, and for each sub-block and causal pixel regions, Histogram of Oriented Gradients (HOG) and Gray-Level Co-occurrence Matrix (GLCM) features (Entropy, Energy, Homogeneity, Correlation, Dissimilarity) are calculated. The combination of all feature sets (i.e. NI, PI, BI, SI) results in a final vector of 115 features representing the CU.

\subsection{RD costs approximation}
\label{subsec:rdcosts}
The goal of both solutions is to approximate RD costs for each splitting mode of an input feature vector. Unlike previous methods that predict split decisions, our approaches focus on predicting normalized RD costs before making split decisions. This allows greater flexibility in determining the split strategy, as differences in RD costs can guide which splits to explore. Unlike classification based methods, regressing RD costs captures smoother transitions in RD cost variations with changes in texture and coding parameters as shown in Figure \ref{fig:VTM-18.0} (Right).
In the regression based RD costs prediction, two cases can be considered. The first case, where the NN is trained on only one CU size, for example $16 \times 16$, the output of this model is the QT cost normalized by the NS cost $(C_{NS})$ of this CU.
The second case, where more than one CU size are considered, the outputs of the NN are two RD costs (NS, QT) normalized by the median of all costs $(C_{median})$ in the dataset. The RL technique follows the second case of regression based method. For a given cost $C$ the normalized cost $(C_{norm})$ is defined as follows:

{\small
\begin{equation}
C_{norm}  = \left\{ \begin{array}{cl}
\frac{C}{C_{NS}} & , \ \text{1 CU size} \\
\frac{C}{C_{median}} & , \text{Otherwise}
\end{array} \right.
\end{equation}
}

\begin{figure}[ht]
    \centering
    \includegraphics[height=5.5cm, width=0.8\textwidth]{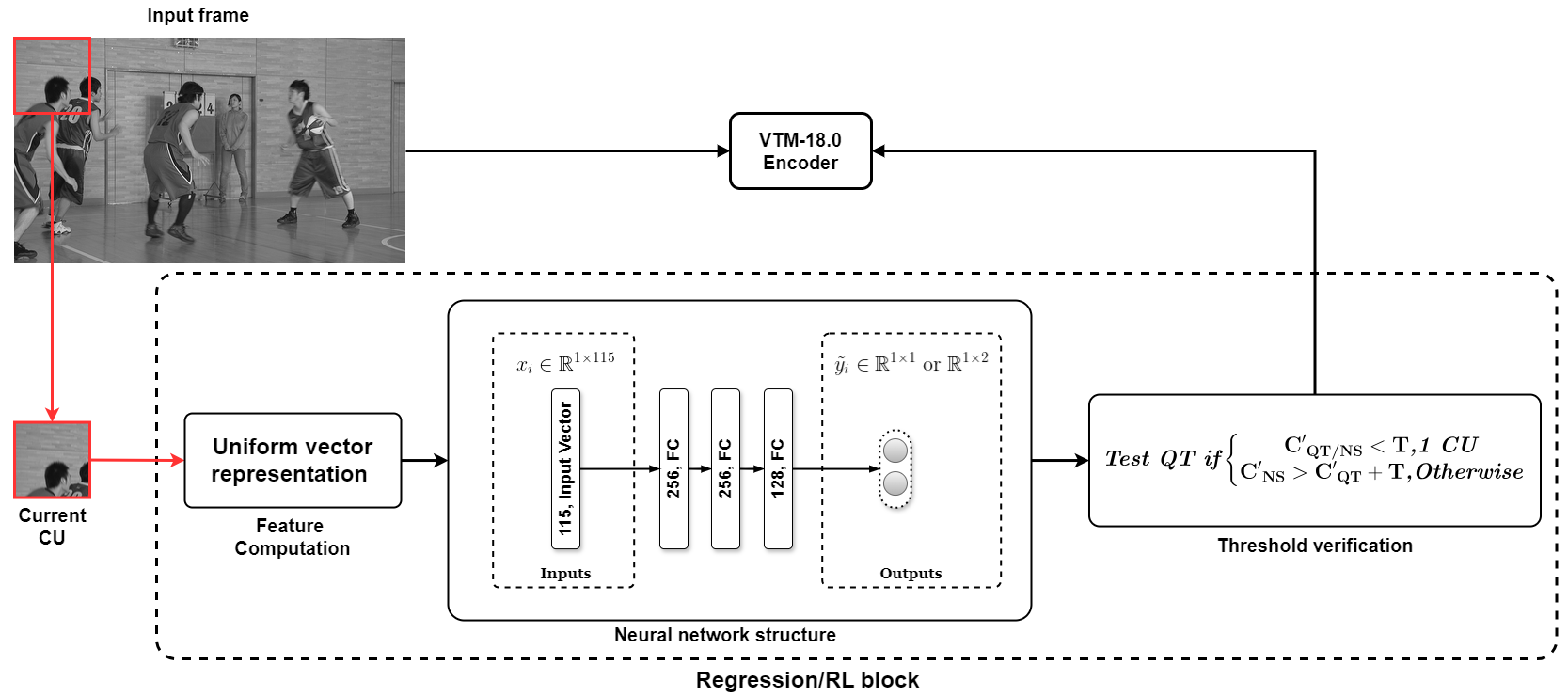}
    \vspace*{-0.7cm}
    \caption{Workflow diagram of the proposed methods}
    \label{fig:RegRLNN}
\end{figure}

\section{Regression based  algorithm overview}

\label{sec:Reg}
The proposed solution predicts normalized RD costs using a uniform vector representation outlined in section \ref{sec:IO}.  As discussed in section \ref{subsec:rdcosts}, the output varies based on the CU sizes processed by a single NN. Fig. \ref{fig:RegRLNN} illustrates the framework architecture. The NN receives a feature vector of size $115$, followed by three fully connected layers with $256$, $256$, and $128$ neurons, respectively. The final output layer may produce either one normalized QT RD cost relative to the NS RD cost or two outputs representing the NS and QT RD costs normalized by the median of all costs.
To build the training dataset, BVI-DVC \cite{BVI-DVC} sequences are encoded using VTM-18.0 in all intra configuration with a maximum QTDepth = 4 and four QP values \{i.e. 22, 27, 32, 37\}. For each CU of size $32\times32, 16\times16$ and $8\times8$ the feature vector is calculated and extracted along with the NS and QT RD costs then a balanced dataset across splitting modes  is generated with 10 million, 20 million and 40 million samples per CU size respectively.
Five NNs were trained using balanced datasets. Three of these networks each with a single output were independently trained on their corresponding datasets, one for each CU size (i.e. $N_{32}$,  $N_{16}$, $N_{8}$). A NN denoted $N_{32,16}$, was trained with two outputs for CUs of sizes $32\times32$ and $16\times16$, using a combined dataset of both CU sizes. The final network, denoted $N_{32, 16, 8}$, was trained with two outputs corresponding to all CU sizes, using a dataset that integrates samples from the $32\times32$ , $16\times16$ and $8\times8$ CU sizes.
The function $f_{CU, \theta}$ defines the NN parameterized by $\theta$ and $CU$ defines the list of the CU sizes treated by this NN where $CU \in \{32, 16,8\}$. The function $f_{CU, \theta}$ takes as input $x_i$, the feature vector of size $115$ and outputs one or two normalized RD costs and is calculated as follows:

{\small
\begin{equation}
  \widetilde{y}_i = f_{CU, \theta}(x_i),
\end{equation}

}

where $\widetilde{y}_i \in \mathbb{R}^{1\times1}$ if a NN is trained on one CU size and $\mathbb{R}^{1\times2}$ otherwise. 

The NNs are trained using Adam optimizer, learning rate of $10^-5$ and a batch size of $512$. The mean square error (MSE) is used as a loss function $\mathcal{L}$ and calculated between the predicted normalized RD costs and the ground truth $y_i$ as follows:

\begin{equation}
\label{eq:loss}
    \mathcal{L}=   \left\| y_i - f_{CU,\theta}(x_i) \right\| 
\end{equation}
The models were trained on 10 epochs in a few hours using a single GPU (NVIDIA Tesla V100 16GB).

\section{RL based algorithm overview}
\label{sec:RL}
DQN can be used to approximate the Q-value function which estimates the expected accumulated reward of taking an action in a particular state. Block partitioning problem can be modeled as a MDP.  In partitioning, the state $S$ represents a given CU which means that the next CU state $S_{i+1}$ relies only on the current CU state $S_i$ and action $A_i$ which is the selected partitioning mode in the RDO process.  In RL, the optimal Q-value function $Q^*(S, A)$ represents the expected cumulative reward when starting in state $s$, taking action $A$, and then following an optimal policy $\pi^*$. The optimal Q-value function is defined as follows:
{\small
\begin{equation}
    Q^*(S, A) = \mathbb{E} \left[ R_t + \gamma \max_{A'} Q^*(S', A') \mid S, A \right]
\end{equation}
}
Where $S$ and $A$ are the current state and action, respectively, $R_t$ is the reward received at time step $t$, $S'$  is the next state after taking action $A$ and $\gamma \in [0, 1]$ is the discount factor, which determines the importance of future rewards.

In DQN, the optimal Q-value function is approximated by a NN parameterized by $\theta$, which takes the state as input and outputs the Q-values for all possible actions and can be denoted as $Q(S, A; \theta)$.

\subsection{DQN training}

Deep NNs can be used as regression model in RL algorithms to learn a Q-function that approximates the RD cost as a reward signal for each action (NS, QT) for two partitioning depths $32\times32$ and $16\times16$ denoted $Q_{32,16}$. The state $S_{l}$ is the uniform vector that represents a $32\times32$ CU in depth $l$, $A_l$ is the action where $A_l \in \{NS, QT\}$  and $\Phi$ is the set of next states $S_{l+1,j}$ of the next depth $l+1$ where $ j \in  [0, 3]$  which represents the four $16\times16$ sub-CUs ($\Phi = \emptyset $ if action on $S_l$ is NS) . The NN is trained on partitioning trajectories of two depths to learn the Q-function. To collect CU partitioning trajectories for the training process, BVI-DVC \cite{BVI-DVC} sequences are encoded using VTM-18.0 in AI configuration with a maximum QTDepth = 4 and four QP values \{i.e. 22, 27, 32, 37\}. During the encoding, for each $32\times32$ CU and its four $16\times16$ sub-CUs the feature vectors are calculated and extracted with the NS and QT RD costs of each CU.  Ending up with 10 million balanced $32\times32$ trajectories (5M for each action).
\begin{algorithm}
\scriptsize
\caption{DQN Algorithm for two depths CU partitioning}
\label{alg:DQN}
\begin{algorithmic}[]
    \State \textbf{Initialize} replay memory \( \mathcal{D} \)
    \State \textbf{Initialize} Q-network with $N_{32,16}$ weights \( \theta \)
    \For{each episode}
        \For{each time step}
            \State Select action \( A_{l,i} \) using an \( \epsilon \)-greedy policy based on \( Q(S_{l,i}, A_{l,i}; \theta) \)
            \State Execute action \( A_{l,i} \), observe reward \( R_{l,i} \) and next state \( \Phi_i \)
            \State Store transition \( (S_{l,i}, A_{l,i}, R_{l,i}, \Phi_i) \) in replay buffer \( \mathcal{D} \)
            \State Sample mini-batch of transitions \( (S_{l,i}, A_{l,i}, R_{l,i}, \Phi_i) \) from \( \mathcal{D} \)
            \State Set \(R_{l,i} \): \[ R_{l,i}  = \left\{ \begin{array}{cl}
            R_{l,i} & , \ \text{if } \Phi_i = \emptyset\text{ or } Depth = l+1 \\
            \Delta_{QT} + \sum_{\Phi_i} min_{A_{l+1, j}} Q(S_{l+1, j}, A_{l+1, j}; \theta) & , \text{Otherwise}
            \end{array} \right.
            \]
            \State Compute the loss:
            \[           \mathcal{L}(\theta) = \mathbb{E} \left[ \left( R_i - Q(S_{l,i}, A_{l,i}; \theta) \right)^2 \right]
            \]
            \State Perform gradient descent step to update \( \theta \)
        \EndFor
    \EndFor
\end{algorithmic}
\end{algorithm}
\medskip

Unlike traditional RL, in the RDO process, the optimal QT RD cost of a CU is the sum of the minimum RD costs of its four sub-CUs, plus additional bits ($\Delta_{QT}$) for coding the QT split and is defined as follows:
{\small
\begin{equation}
\label{AGRCosts}
    C_{QT}^l = \sum_{i=0}^{3} \min( C_{NS, QT}^{(i, l+1)}) + \Delta_{QT}
\end{equation}
}
Where $C_{QT}^l$ is the QT RD cost of a CU at a depth $l$, $\min( C_{NS, QT}^{(i, l+1)})$  is the minimum RD cost of the $i$-$th$ sub-CU in the next depth and $\Delta_{QT}$ is the RD cost of signaling the QT split of the current CU. In the training phase, the Eq. (\ref{AGRCosts})  is modified to calculate the target $32\times32$ QT RD cost based on the cost of the action (NS or QT) actually selected by the agent in the next depth (i.e. $16\times 16$ CUs) and $\Delta_{QT}$ remains as the original value. This enables the agent to understand the relationship between depths and maintain consistent behavior, accounting for suboptimal choices at lower levels. The targets for NS costs and QT costs for $16\times16$ remain as original values, with all target costs normalized by the median cost of the entire training dataset. A replay memory \( \mathcal{D} \) is used to store the observed transitions \( (S_{l,i}, A_{l,i}, R_{l,i}, \Phi_i)\) and to select action on a given state, \( \epsilon \)-greedy strategy is applied. 
The resulting agent approximates RD costs of splitting modes for $32\times32$ and $16\times16$ blocks. The action can be selected based on a pre-determined threshold for a better trade-off. Algorithm \ref{alg:DQN} summarizes the training steps.

The DQN and regression methods were trained offline in this study, despite the DQN algorithm being ideal for online training, which allows dynamic adaptation to various VTM encoding configurations. Although training during encoding is feasible, it would add significant computational overhead. The regression method relies on static datasets for specific configurations, while DQN provides greater robustness through real-time learning. The models of both solutions are deployed as SADL \cite{SADL2021} models for integration within VTM for inference

\begin{figure}[ht]
    \centering
    \includegraphics[height= 10cm, width=0.75\textwidth]{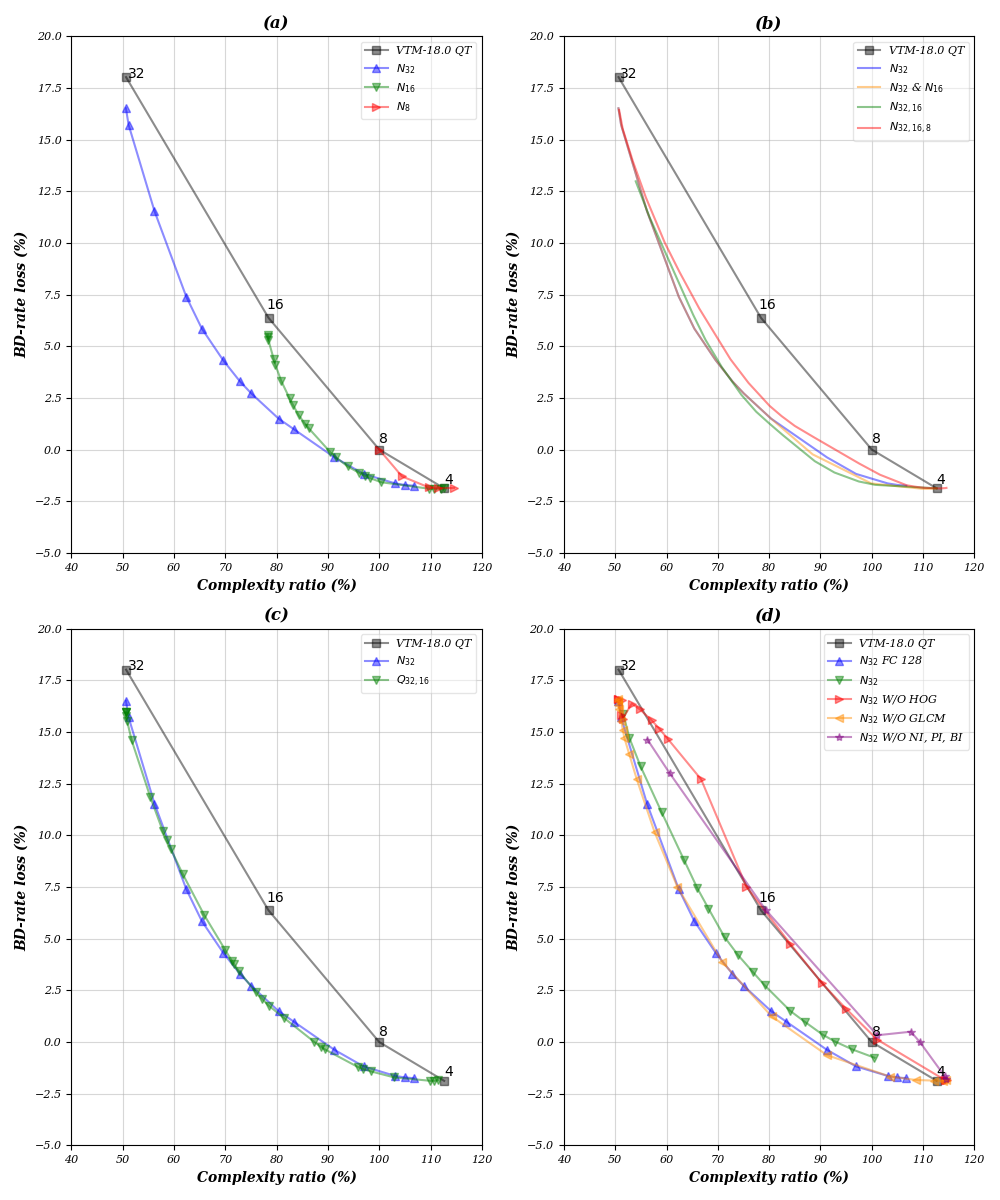}
    \vspace*{-0.2cm}
    \caption{BD-rate versus complexity speed-up ratio comparison in All Intra (AI) configuration with four QTDepth configurations.}
    \label{fig:Results}
\end{figure}

\section{Experimental results}
\label{sec:Exp}
Both solutions are integrated into VTM-18.0 and compared with the benchmark curve of VTM-18.0 discussed in Section \ref{sec:VTMQT}. Test sequences from the JVET Common Test Conditions \cite{jvet:ctc} are used for evaluation. Fig. \ref{fig:Results} presents the complexity-gains trade-offs achieved by both approaches compared to the VTM-18.0 reference benchmark.

\subsection{Regression NNs versus DQN algorithm}
\label{subsec:RegRL}
During inference, both solutions predict normalized RD costs. NNs that predict a single value, denoted as $C'_{QT/NS}$, represent the predicted normalized QT RD cost per NS RD cost. For the other NNs, the two predicted values are $C'_{NS}$ and $C'_{QT}$. A grid search is used for pre-selecting thresholds over the dataset, which is then applied to the test sequences. For each threshold $T$, the associated bd-rate loss and speed-up trade-offs are obtained, as outlined in Fig. \ref{fig:RegRLNN}. 
Fig. \ref{fig:Results} (a) shows regression based complexity-gains trade-offs: $N_{32}$ (blue), $N_{16}$ (green), and  $N_{8}$ (red) are tested individually against the anchor. $N_{8}$ can accelerate partitioning from QTDepth 4 to 3 without bd-rate loss. A lossless speed-up of 10\% is possible with $N_{32}$ and $N_{16}$. $N_{32}$ adds superior trade-offs compared to $N_{16}$ and the anchor, for example a $25\%$ speed-up at a bd-rate loss of $2.70\%$.
Fig. \ref{fig:Results} (b) shows trade-offs for $N_{32,16}$, $N_{32,16,8}$ and combined $N_{32}$, $N_{16}$. $N_{32,16}$ offers better trade-offs up to $30\%$  complexity reduction compared to $N_{32}$ alone. Then $N_{32}$ takes over due to its specialization in $32\times32$ CUs, providing better performance than $N_{32,16}$. When both $N_{32}$, $N_{16}$ are activated, they offer superior trade-offs compared to  $N_{32}$ alone, as $N_{16}$ adds acceleration. After $20\%$ of complexity reduction, both scenarios show similar trade-offs. $N_{32,16,8}$ accelerates partitioning for all CUs (i.e. $32\times32$, $16\times16$ and $8\times8$) but requires extensive threshold search, whereas $N_{32,16}$ uses a simpler grid search.
Fig. \ref{fig:Results}  (c) shows the $Q_{32,16}$ curve alongside $N_{32}$. The RL model behaves similarly to $N_{32,16}$ but surpasses $N_{32, 16 }$ after $35\%$ of complexity reduction. 
Table \ref{tab:ctcseq} shows a comparison with VTM-18.0. $N_{32}$ achieves a $19.61\%$ complexity reduction with a $1.495\%$ bd-rate loss while $N_{32,16}$ and $Q_{32,16}$ offer trade-offs of  $20.15\%$, $1.302\%$ and $18.63\%, 1.18\%$ respectively. For example in the class D sequences, $Q_{32,16}$ achieves an $11\%$ of complexity with a minimal bd-rate loss of $0.103\%$. $N_{32,16}$ outperforms $Q_{32,16}$ in complexity reduction by $1.7\%$ but with a higher bd-rate increase of $0.384\%$. In contrast, $N_{32}$ offers an $7.78\%$ reduction in complexity with bd-rate loss of $0.198\%$ indicating that $Q_{32,16}$ achieves $3.3\%$ greater complexity reduction while maintaining a bd-rate advantage of $0.095\%$.

\begin{table}[ht]
\centering

\caption{$\Delta$C and bd-rate trade-offs for $N_{32}$, $N_{32,16}$ and $Q_{32,16}$ in AI coding configuration compared to VTM-18.0}
\scriptsize
\begin{tabular}{cccccccc}
\hline
\multicolumn{1}{c}{{}} & {} & \multicolumn{2}{c}{$N_{32}$} & \multicolumn{2}{c}{$N_{32,16}$} & \multicolumn{2}{c}{$Q_{32,16}$} \\ \hline

\multicolumn{1}{c}{{ Class}} & { Sequence} & { bd-rate\%} & { $\Delta$C\%} & { bd-rate\%} & { $\Delta$C\%} & { bd-rate\%} & { $\Delta$C\%} \\ \hline

\multicolumn{1}{c}{{ }} & { Tango2} &{ 3.166} & { 64.87} & {1.941} & {69.4} & {2.312} & {70.53} \\

\multicolumn{1}{c}{{ }} & { FoodMarket4} &{ 1.407} & { 71.36} & {1.496} & {72.4} & {1.766} & {71.41} \\

\multicolumn{1}{c}{\multirow{-3}{*}{{ Class A1}}} &{ Campfire} &{ 1.136} &{ 80.54} & {0.945} & {80.98} & {1.09} & {82.2} \\ \hline

\multicolumn{2}{r}{{ \textbf{Average}}} & { \textbf{1.903}} & \textbf{72.26} & \textbf{1.46} & \textbf{74.26} & \textbf{1.722} & \textbf{74.71}  \\ \hline

\multicolumn{1}{c}{{ }} & { CatRobot1} & { 2.713} & { 75.1} & {2.339} & {75.52} & {2.188} & {77.53} \\

\multicolumn{1}{c}{{ }} & { DaylightRoad2} & { 4.54} & { 69.94} & {1.828} & {75.25} & {0.784} & {78.61} \\

\multicolumn{1}{c}{\multirow{-3}{*}{{ Class A2}}} & { ParkRunning3} &{ 0.609} &  {71.28} & {0.744} & {71.91} & {0.869} & {72.34} \\ \hline

\multicolumn{2}{r}{{ \textbf{Average}}} &{ \textbf{2.62}} & \textbf{72.10} & \textbf{1.637} & \textbf{74.22} & \textbf{1.278} & \textbf{76.16} \\ \hline

\multicolumn{1}{c}{{ }} &{ MarketPlace} &{ 2.207} & {64.2} & {1.64} & {70.06} & {1.516} & {73.93} \\

\multicolumn{1}{c}{{ }} &{ RitualDance} &{ 3.413} & {75.6} & {3.349} & {74.34} & {5.489} & {72.38} \\

\multicolumn{1}{c}{{ }} &{ Cactus} & { 1.411} & {76.7} & {1.36} & {78.44} & {0.789} & {82.94} \\

\multicolumn{1}{c}{{ }} &{ BasketballDrive} & { 3.015} & { 60.82} & {1.943} & {66.57} & {1.847} & {69.15}  \\

\multicolumn{1}{c}{\multirow{-5}{*}{{ Class B}}} &{ BQTerrace} &{ 0.9} &{84.67} & {0.705} & {85.35} & {0.390} & {86.68} \\ \hline

\multicolumn{2}{r}{{ \textbf{Average}}} &  { \textbf{2.19}} & \textbf{72.4} & \textbf{1.8} & \textbf{74.95} & \textbf{2} & \textbf{77.02} \\ \hline

\multicolumn{1}{c}{{ }} &{ RaceHorses} & { 0.367} & { 88.34} & {0.732} & {84.39} & {0.417} & { 85.62}  \\

\multicolumn{1}{c}{{ }} &{ BQMall} & {1.016} & {88.13} & {1.205} & {83.81} & {0.959} & {84.4} \\

\multicolumn{1}{c}{{ }} &{ PartyScene} & { 0.341} & {97.76} & {0.296} & {94.29} & {0.026} & {96.79} \\

\multicolumn{1}{c}{\multirow{-4}{*}{{ Class C}}} & { BasketballDrill} &{ 1.37} &{89.21} & {2.299} & {85.6} & {1.563} & {85.9} \\ \hline

\multicolumn{2}{r}{{ \textbf{Average}}} & \textbf{0.774} &  \textbf{90.86} & \textbf{1.133} & \textbf{87.022} & \textbf{0.741} & \textbf{88.17} \\ \hline

\multicolumn{1}{c}{{ }} &{ RaceHorses} & { -0.059} & {95.66} & {0.559} & {89.56} & {-0.254} & {91.66} \\

\multicolumn{1}{c}{{ }} & { BQSquare} & { 0.207} & {95.33} & {0.174} & {93.68} & {-0.214} & {95.15} \\

\multicolumn{1}{c}{{ }} & { BlowingBubbles} & { -0.198} & {97.59} & {0.271} & {89.09} & {-0.008} & {91.92} \\

\multicolumn{1}{c}{\multirow{-4}{*}{{ Class D}}} &{ BasketballPass} &{ 0.842} &{80.31} & {0.942} & {76.52} & {0.889} & {76.86} \\ \hline

\multicolumn{2}{r}{{ \textbf{Average}}} & \textbf{0,198} & \textbf{92.22} & \textbf{0.487} & \textbf{87.21} & \textbf{0.103} & \textbf{88.9} \\ \hline

\multicolumn{2}{r}{{ \textbf{Average A-B-C-D}}} & \textbf{1.495} & \textbf{80.39} & \textbf{1.302} & \textbf{79.85} & \textbf{1.18} & \textbf{81.37} \\ \hline

\end{tabular}
\label{tab:ctcseq}
\end{table}
\medskip
\vspace*{-0.45cm}

\begin{table}[ht]
\centering
\caption{Impact of feature sets and model complexity on complexity-gains trade-offs.}
\scriptsize
\begin{tabular}{l|c|c|c|c|c}
\hline
\multirow{2}{*}{\textbf{Feature Set / Model}} & \multicolumn{3}{c|}{\textbf{Overall Impact Level}} & \multicolumn{2}{c}{\textbf{bd-rate \%}} \\ \cline{2-6} 
 & Low & Moderate & High &  \textbf{ $\Delta$C = 10\%} & \textbf{ $\Delta$C = 20\%} \\ \hline

\textbf{W/O NI, PI, BI}  &   -  & - & \checkmark & 3.301 &  6.405 \\ 
\textbf{W/O HOG}        & - & -  &    \checkmark & 2.853  & 6.375 \\ 
\textbf{W/O GLCM} & \checkmark &   -  &   -   &  -0.53 & 1.245 \\ 
\textbf{Reduced Model }& \checkmark & \checkmark &  -  & 0.355  &  2.115    \\ 
\textbf{Proposed N$_{32}$}   & - & - &  -  &  -0.379 &   1.495   \\\hline
\end{tabular}

\label{tab:ablation_study}
\end{table}
\medskip
\vspace*{-0.45cm}

\subsection{Ablation study}
The ablation study evaluates the contribution of each input feature set to the overall performance for $32\times32$ CUs. $N_{32}$ incorporate all feature sets, and specific features were systematically removed for retraining.  Additionally, the model complexity is also evaluated where the number of neurons in each fully connected layer was reduced from $256, 256, 128$ to $128, 128, 64$, respectively. Fig. \ref{fig:Results} (d) and Table \ref{tab:ablation_study} show the performance variations across different feature combinations along with the overall impact level.

\section{Conclusion}
\label{sec:Conc}
Two main techniques are proposed and compared to accelerate the intra partitioning process based on a vector of fixed dimension. This allows the two methods to be size independent and can treat more than one CU size. The methods are a regression based NN and RL agent that estimate the RD costs of a given CU. The RL algorithm is adapted to take in consideration the predictions in the next depth to give the agent more understanding. In this study, the proposed models are only used to compute split decision. In a future work, the use of the prediction of the relative RD cost will be used for other tasks such as rate control.

\Section{References}
\bibliographystyle{IEEEbib}
\bibliography{refs}

\end{document}